\documentclass[a4paper]{article}

\usepackage{INTERSPEECH2020}
\usepackage{url}
\usepackage{subcaption}
\usepackage{xcolor}

\title{Insightful Assistant: AI-compatible Operation Graph Representations for\\Enhancing Industrial Conversational Agents}

\name{Bekir Bayrak, Florian Giger, Christian Meurisch}

\address{Technical University of Darmstadt, Germany}
\email{\{bayrak,giger,meurisch\}@tk.tu-darmstadt.de}

\begin{document}

\maketitle

%-abstract
\begin{abstract}
Advances in voice-controlled assistants paved the way into the consumer market.
For professional or industrial use, the capabilities of such assistants are too limited or too time-consuming to implement due to the higher complexity of data, possible AI-based operations, and requests.
In the light of these deficits, this paper presents \textit{Insightful Assistant}---a pipeline concept based on a novel operation graph representation resulting from the intents detected.
Using a predefined set of semantically annotated (executable) functions, each node of the operation graph is assigned to a function for execution.
Besides basic operations, such functions can contain artificial intelligence (AI) based operations (e.g.,~anomaly detection).
The result is then visualized to the user according to type and extracted user preferences in an automated way.
We further collected a unique crowd-sourced set of 869~requests, each with four different variants expected visualization, for an industrial dataset.
The evaluation of our proof-of-concept prototype on this dataset shows its feasibility: it achieves an accuracy of up to 95.0\% (74.5\%) for simple (complex) request detection with different variants and a top3-accuracy up to 95.4\% for data-/user-adaptive visualization.

\end{abstract}

%-terms

%-all sections
\section{Introduction}

%-motivation
In the last decade, digital assistants have made the transition from research prototypes to consumer products, entering our lives in the form of Apple Siri~(2011) or Amazon Alexa~(2015), to name just a few~\cite{luger2016like}.
With such voice-controlled systems (also referred to as \textit{conversional agents} [CAs]~\cite{grudin2019chatbots}), users can request (mostly general) information and have simple tasks (e.g.,~making a calendar entry) performed~\cite{sarikaya2017technology}.
While the current capabilities of conversational agents are assumed in the consumer market, their use in professional or industrial context tends to stagnate due to the higher complexity of data, AI-based data operations (e.g.,~anomaly detection), and requests~\cite{xu2019big,kandel2012enterprise}.
In other words, there is an indispensable need for digital assistants that understand domain-specific requests for business data as well as process, prepare, and visualize them accordingly.

%-example
To illustrate this complexity in an industrial context, the following requests are given as an example.
``Show me the average power consumption of all machines in Hall 12 built after 2017'' is a simple request, as it can be mapped to database queries and operations -- the visualization can be a plot, a table, or even both to provide the user with all relevant information.
In contrast, the request ``which machine in Hall 12 will fail next?'' is more complex, as it requires the use of AI-based algorithms (e.g.,~from the discipline of predictive maintenance) -- but the visualization is somewhat simpler, which can be a voice output, a textual machine description, or both depending on the user's preferences.

\begin{figure}[t]
  \includegraphics[width=1\columnwidth]{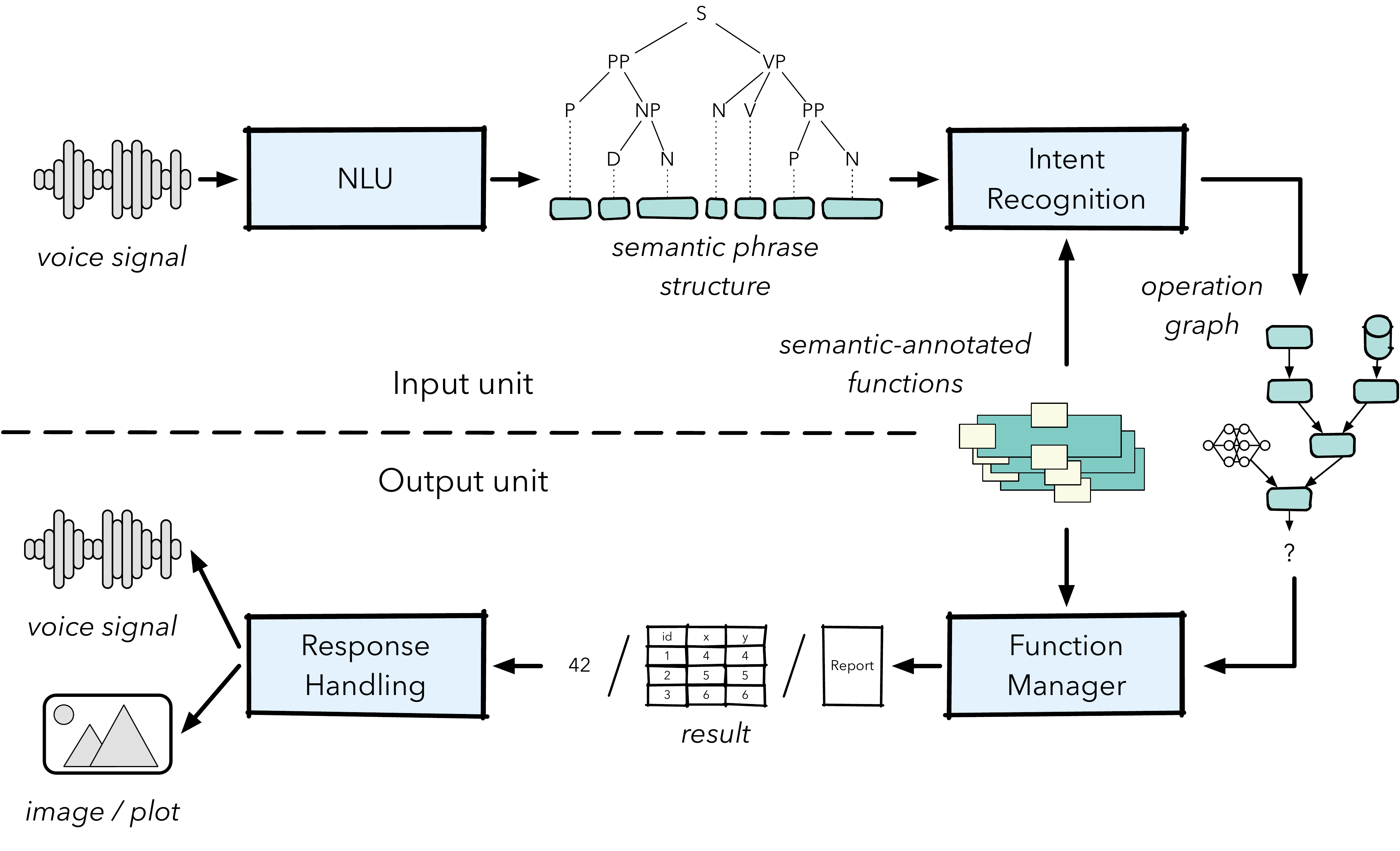}
  \caption{Pipeline of our proposed Insightful Assistant, comprising three components: an input unit (semantic understanding), the AI-compatible operation graph representation in the core, and an output unit (data-/user-adaptive visualization).}
  \label{fig:pipeline}
\end{figure}

%-related work
While latest approaches already achieve accurate and robust speech recognition~\cite{sarikaya2017technology,tur2008calo} to identify user intents~\cite{sun2016contextual,zhang2017intent,bhargava2013easy} using semantic parsing~\cite{marzinotto2019adapting,bhat2019neural}, the mapping to advanced data operations (AI algorithms) is limited -- and their extensibility is challenging, requiring a high effort.
Most often in this context, intents are mapped to SQL query language of databases.
Examples include \textit{Seq2SQL}~\cite{zhong2017seq2sql} and \textit{SpeakSQL}~\cite{chandarana2017speakql} -- these approaches do \textit{not} make the integration of AI algorithms straightforward.
Looking at the output, the results are mostly returned as voice output~\cite{jaimes2007multimodal}.
However, data-exploring requests often require a richer, cross-modal output~\cite{huang2019detecting}. 
Examples that can provide such output include \textit{DeepEye}~\cite{luo2018deepeye}, \textit{Data2Vis}~\cite{dibia2019data2vis}, \textit{D3}~\cite{bostock2011d3}, \textit{VizML}~\cite{hu2019vizml}, and \textit{CompassQL}~\cite{wongsuphasawat2016towards} -- but they are neither integrated in CAs nor do the internal representations provide the corresponding information to the output unit.

%-in this paper (concept)
In the light of these deficits, this paper presents \textit{Insightful Assistant}---a pipeline concept based on a novel operation graph representation that enables better integration of AI-based algorithms for more advanced data analysis and visualization (see Figure~\ref{fig:pipeline}).
In particular, an adapted input unit maps the user intents recognized from the semantically analyzed voice input to this new representation.
The operation graph is then resolved by assigning its nodes to (executable and generic) functions from a predefined set with semantic annotations -- such functions can contain `basic' operations (e.g.,~summation or averaging) as well as AI-based algorithms/models (e.g.,~anomaly detection), which can ad hoc analyze the data in depth.
If present in the intent, this representation also contains output information: 
an adapted output unit selects a suitable result presentation depending on the type of the result and the user's preferences.

%-in this paper (evaluation)
As there is \textit{no} dataset of requests with corresponding ground-truth data in an industrial context to evaluate our operation graph approach, we have further collected them using a crowd-sourcing approach and an underlying honey production dataset.
The resulting dataset contains 3,476 request variants (869 requests \`a 4 variants having the same user intent), ranging from simple to complex questions.
The evaluation shows accurate results, demonstrating the feasibility of \textit{Insightful Assistant}.

%-contributions
In summary, the contribution of this paper is threefold.
\textit{First}, we propose a novel operation graph representation generating from the intents detected; our operation graph enables processing with AI-based operations and data-/user-adapted visualization (see Section~\ref{sec:concept}).
\textit{Second}, we collected a unique crowd-source set of requests for a given industrial dataset; each request has four different variants, ranging from simple to complex (see Section~\ref{sec:dataset}).
\textit{Third}, we evaluate a prototypical implementation of \textit{Insightful Assistant} on the collected set of requests, showing its feasibility: it further achieves an accuracy up to 95\% and 75\% for simple and complex requests, respectively (see Section~\ref{sec:eval}).

\section{Background \& Related Work}

As the contributions do \textit{not} lie on speech recognition but rather on the internal representation (operation graph generation) from intent recognition as well as its processing and result visualization, we give an overview of the related work only for them in the following -- related surveys are given in \cite{sarikaya2017technology,de2020intelligent}.

%-intent recognition/ internal representation
As first part (\textit{input}), conversational agents aim to extract the meaning from speech utterances (e.g.,~CALO~\cite{tur2008calo}).
To this end, latest approaches rely on a so-called NLU (natural language understanding) component to take the speech transcription~\cite{sarikaya2014application,tur2011spoken} and perform the semantic analysis~\cite{deoras2012joint,kate2005learning} for determining the underlying user intent~\cite{liu2015deep,sun2016contextual}.
This intent then retrieves its answer candidate from various resources, such as web documents, search engines, Wikipedia, or a \textit{semantic knowledge graph} (KG)---``representing information using triples of the form subject-predicate-object where in graph form the predicate is an edge linking an entity (the subject) to its attributes or another related entity''~\cite{ma2015knowledge,sarikaya2017technology}.
Examples include Freebase, Facebook's Open Graph, Thingpedia, and DBpedia~\cite{campagna2017almond,bobadilla2013recommender}.
Despite multi-domain support~\cite{khan2015hypotheses,el2014extending}, in the industrial context for ad-hoc data analysis, these approaches are rather complex or difficult to apply.
As a different approach, the intent can also be translated into \textit{structured query language} (SQL) for retrieving information from an underlying database, which is often the framework in an industrial context.
Examples include SpeakSQL~\cite{chandarana2017speakql} and seq2sql~\cite{zhong2017seq2sql}.
While the former relies on keyword matching for filling the placeholders in SQL statements, the latter proposes an AI-based approach: it uses a deep neural network for performing this translation.
Both internal representations (KG, SQL) are capable of retrieving (`simple') requested information; however, they are limited in their ability to integrate AI-based operations and ad-hoc data analysis.

%-visualization
As second and last part (\textit{output}), conversational agents aim to present their results to the user accordingly.
Most conversational agents translate the machine-readable result back to natural language; they further use the same output modality as the input modality (voice), supported by written text~\cite{sarikaya2017technology}.
While this approach works well in the consumer market, it is very limited in the industrial context when presenting the results of (business or production) data analysis.
Here, in particular, richer and \textit{data-adaptive visualization} methods are required.
Examples for such methods include DeepEye~\cite{luo2018deepeye}, Data2Vis~\cite{dibia2019data2vis}, D3~\cite{bostock2011d3}, VizML~\cite{hu2019vizml}, and CompassQL~\cite{wongsuphasawat2016towards}.
However, only few conversational agents integrate such methods in industrial context -- an ad-hoc adaptation to the users' preferences is also still missing.
\section{Insightful Assistant Design}
\label{sec:concept}

In this section, we present the pipeline of \textit{Insightful Assistant}---comprising an adapted input unit, the novel operation graph representation as the interface, and an adapted output unit (see Figure~\ref{fig:pipeline}).
The operation graph connects both, the input unit and the output unit, aiming to enable a better integration of AI algorithms for data retrieval and a richer, user-adapted output.

%-input unit
\subsection{Input Unit: Semantic Understanding}
The pipeline starts with the input unit to semantically understand the user's spoken request and recognize the user's intent.
The former is achieved by the \textit{natural language understanding} (NLU) component---comprising the \textit{automatic speech recognition} (ASR) and the \textit{natural language processing (NLP) annotation}.
The ASR component translates the voice input (speech) to text, which is already a very well-researched topic.
Next, the NLP annotation component applies tokenization, dependency parsing, and part-of-speech (POS) tagging -- we use \textit{spaCy}\footnote{https://spacy.io} to implement this component.
Its result is a \textit{semantic phrase structure}, which prepares our contributions below.

%-operation graph
\subsection{The Core: Operation Graph Representation}
The latter part of the input unit---namely our \textit{intent recognition} component---analyzes this phrase structure to assign parts of them to predefined semantic-annotated functions $F$.
A function $f_j \in F$ has an input $a_j$ (data and parameters), an output $o_{j} = f_{j}(a_j)$, and a semantic description $s_j$; it can contains multiple data operations $\{c_i|i=1..k\}$, representing a sub operation graph.
Data operations range from `basic' operations (e.g.,~summation, averaging, or filtering) to AI-based algorithms/models (e.g.,~anomaly detection), which can ad hoc analyze the data in depth~\cite{meurisch2019assistantgraph}.
The semantic linking of these functions by the phrase structure creates the whole operation graph representation (see below).
Formally, let's consider an operation graph as a directed acyclic graph $G = (C, E)$ with a set of atomic operations (nodes) $C = \{c_{1},...,c_{n}\}$ and dependencies (edges) $E = \{e_{1},...,e_{m}\}$ between these operations (m, n may vary with different operation graphs). 
Each dependency edge $e_{q \rightarrow p} \in E$ is a tuple $e_{q \rightarrow p} = \{c_q, c_p\}$, where the input type of $e_p$ must correspond to the output type $e_q$, i.e.,~$a_p = o_q$.

To build this operation graph automatically, the semantic phrase structure is recursively resolved:
(i)~the given part of this structure must `match' the semantic description of the possible (executable and generic) functions -- keyword matching, word embeddings, and synonyms are used for this~\cite{liu2015topical};
(ii)~keys/references to the underlying data sets or databases are then identified to retrieve the required data later -- again, keyword matching, word embeddings, and synonyms are used for this;
(iii)~for each possible function the recursion is continued, creating multiple operation graphs; the output of the current function limits the set of possible functions (the input type must match this output type) in the next iteration.
Finally, if no complete operation graph can be generated, the one with the greatest depth is selected; if there is more than one operation graph, the most relevant and most appropriate one is selected; otherwise, only the one created is selected.

Next, the \textit{function manager} handles the orderly calling of the required operations of $G$ in the execution environment of~\cite{meurisch2019assistantgraph}, passing parameters and forwarding the results.
In case the execution environment runs locally (e.g.,~at customer site) or on third-party devices, appropriate mechanisms such as in ~\cite{brasser2018voiceguard,meurisch2020privacy} can be integrated to protect both the intellectual property~(IP)---e.g.,~for the underlying AI models of the predefined functions---and the business data.
The result of this execution is the \textit{machine-readable answer} to the given user request.

%better integration for ad-hoc data analysis~\cite{meurisch2019assistantgraph}

%-output unit
\subsection{Output Unit: Data-/User-adaptive Visualization}
The last part of the pipeline is the output unit for the translation of the machine-readable results into visualizations for users -- the \textit{response handling} is responsible for this.
As the type of visualization varies with the data to be visualized and the user's preferences, we have implemented two adaptation mechanisms: \textit{data-adaptive visualization} and \textit{user-adaptive visualization}.

The former relies on historical data $N$. We use k-Nearest Neighbor (kNN) with $k=\sqrt{|N|}$ and majority voting (as kNN offers transparency to determine the top3 visualization results) to train a model that adapts the visualization depending on the resulting data type.
To get a rich visualization, \textit{Insightful Assistant} supports various types including location, numerical, temporal, and categorical values -- or any combination of them (e.g.,~filtered table entries).
According to the data type, appropriate (and compatible) visualization forms are then learned.
For instance, a geographical heat map can be visualized (as one of many forms), when users ask for \textit{``show me the average production output in our plants''} or \textit{``where is the plant with the highest production output located?''}.

The latter additionally takes into account the output information requested by the user that has been recognized in the user intent (if specified) and stored into the internal representation.
Analogous to the data analysis functions from the previous subsection, visualization forms (considering possible input and output parameters) are added by semantic triggers.

Last but not least, our novel operation graph representation further enables a new kind of \textit{data exploration}: 
users can further explore the data by going steps/nodes back in the operation graph; 
the results of these steps are then visualized together (e.g.,~by means of an overlay).
For instance, users can view the underlying values of an averaged output when they go one step (in this case: the `averaging' operation) back.

\section{Dataset}
\label{sec:dataset}

As there is \textit{no} appropriate dataset of requests with corresponding ground-truth data to evaluate our operation graph approach, we have collected one especially for this purpose.
In particular, this section first describes the collection process and the data cleaning process for getting the resulting high-quality dataset.

\subsection{Data Collection}

%0--methodology+recruitment
For the data collection, we use a crowdsourcing approach: 
as a manageable task (a so-called \textit{microtask}), we present an industrial production dataset to users (or micro-workers) and asked them to think up requests that would interest them.
Specifically, we use \textit{Prolific}\footnote{https://prolific.ac}---a crowd working platform for social science experiments~\cite{palan2018prolific}---for the recruitment of participants, \textit{LimeSurvey}\footnote{https://www.limesurvey.org} for the creation of the questionnaire, and a \textit{honey production dataset}\footnote{https://www.kaggle.com/jessicali9530/honey-production} as data basis.
Although this microtask is not limited to a specific group of people, we used the pre-screening feature of Prolific to filter participants by language (English) and education (A level graduation or higher).
Monetary incentives motivated the participants to take part in this questionnaire: 
participants who completed the microtask successfully---how we assessed this is given below---were rewarded with \$8 each.

%1--description
Next, we describe the procedure of the microtask.
First, participants get a detailed description of the microtask; 
they also gain insight into the above dataset and familiarize themselves by simply playing around with the data and answering simple training questions.
%2--questions
Then, we ask for two different types of requests, which we refer to as \textit{simple} and \textit{complex} requests.
For each type of request, we described in detail how they should create the requests, and we gave them sample requests and variants -- the latter expresses the same user intent und thus has the result as the former.
For instance, \textit{``How much was honey in Alabama in 2010?''}, \textit{``What did honey cost in Alabama ten years ago?''}, and \textit{``Show me the average price of honey in AL in 2010''} can be variants for the base request `\textit{`What was the price of honey in Alabama in 2010?''} -- in all cases, the result is \$2.4 (average per pound) and can be simply looked up in the table. 

%2a--simple questions
For each type of request, participants should enter six base requests, each with four different variants in total (2$\times$6$\times$4 = 48~request variants).
For simple requests, we limit the participants to lookup and simple table calculation functions, such as summation, filtering, averaging, to name a few.
Participants were also asked to enter the result of their base request.
%2b--complex questions
For complex requests, we additionally provide the participants with a predefined set of more complex (AI-based) functions with a detailed description of how they work and what they can do.
Among the most sophisticated examples are the anomaly detection and regression analysis modules.
Participants were asked to integrate \textit{at least} one of these AI-based functions/algorithms; simple table calculation functions can still be used additionally.
For instance, a complex request can be the following: \textit{``How will the average honey price develop in Florida next year?''} -- in this case, after filtering the state, a regression model trained from the historical values of the dataset is applied.
Here, the participants were also asked to enter only the expected type of result (not the exact result, which would also not be possible).

%3--visualization
For each request, the participants were also asked to select a suitable type of output, what they would expect (based on their preferences and the type of the result).
We have prepared nine different types of visualization with detailed descriptions -- these include simple text responses, tables, different types of diagrams, (geographical) heat map, to name a few.

%4--plausibility checks
\subsection{Data Quality Assurance \& Data Cleaning}
After the participants have submitted the questionnaire and before they receive their reward, we have manually checked the requests for plausibility.
Meaningless and useless requests or those that do not meet the above conditions have been rejected, and the participant was asked to correct them; 
otherwise, the entire microtask was not accepted, and the participant was rejected completely and thus not rewarded.

After a microtask has been completed, two researchers have independently performed the following tasks for each request to ensure the quality of the dataset:
spelling and grammatical errors were corrected without changing the user intent;
the user-entered results for the simple requests were cross-checked;
for the complex requests, the operation graph is created manually and executed to get the corresponding result;
the results (especially more complex ones such as filtered tables) are then translated into a machine-readable format for automated evaluation.
Discrepancies between the two researchers were discussed in the end, and a decision was made in favor of one version.

Finally, a total of 79~participants have successfully completed the microtask.
The resulting dataset contains \textit{3,476 request variants} (869 requests \`a four variants with the same user intent), ranging from simple to complex questions (fifty-fifty).

\section{Evaluation}
\label{sec:eval}
In this section, we evaluate a prototypical implementation of our proposed concept---starting by describing the methodology.

%-methodology
\subsection{Methodology}

%--assumptions
For our evaluation, we assume a perfect \textit{speech-to-text} unit. 
The reason for this is that this topic is well-researched~\cite{sarikaya2017technology}, and we do contribute to it, relying on an open-source solution.
This means we start with a text input, which is compatible with the collected dataset.
On the output side, we also limit the evaluation to our contributions, the generation of the operation graph and a data-/user-adapted visualization.
This means that we take the machine-readable outputs for an automated evaluation.
More precisely, we take the result of the operation graph (directly after the `function manager') to evaluate its correct creation and execution;
only in case of correct result, we take the visualization command (directly after the `response handling', just before data is visualized) to evaluate the resulting visualization against the user expectations collected for a given request.

%---dataset, baselines
Further, we use our specially collected data described above (see Section~\ref{sec:dataset}).
As baseline approaches for the data-adapted visualization, we integrate majority voting (\textit{ZeroR}), which only takes the majority visualization, and \textit{CompassQL}~\cite{wongsuphasawat2016towards}.

%-results
\subsection{Results}
We now report and discuss the results of our evaluation with respect to the generation of (i)~the right operation graph from text input and (ii)~a suitable data- and user-adapted visualization.

\subsubsection{Accurate Generation of the Operation Graph}
In the first experiment, we investigate how well the system generates and executes the core operation graph.
For simple requests, \textit{Insightful Assistant} achieves a very high accuracy of $95.0\pm0.8\%$ -- it still achieves a high accuracy of $87.5\pm0.7\%$, if different variants are considered.
For complex requests, the accuracy of our proof-of-concept prototype is good ($74.5\pm1.5\%$) -- taking into account variants, moderate accuracy values of $63.6\pm1.3\%$ are still achieved.
Nonetheless, \textit{Insightful Assistant} automatically integrates AI-based operations that would otherwise have to be implemented with increased effort (e.g.,~using keyword matching and static operation graphs).

%TODO: 2 plots (simple/complex) -- number of correct variants
%In addition, \textit{Insightful Assistant} is able to process very different variants of the same user intent, which then logically lead to the same result.
%\myworries{To investigate this in detail, Figure~\ref{fig:variantResults} shows ...}

%\begin{figure}[t]
%	\centering
%    \begin{minipage}{0.47\linewidth}
%    	\centering
%		\includegraphics[width=4.5cm]{figures/plots/densitysimple}
%		\subcaption{Simple questions}
%		\label{subfig:simpleVariants}
%    \end{minipage}
%    \quad
%    \begin{minipage}{0.47\linewidth}
%       \centering
%	   \includegraphics[width=4.5cm]{figures/plots/densitycomplex}
% 	   \subcaption{Complex questions}
%  	   \label{subfig:complexVariants}
%    \end{minipage}
%    \caption{\myworries{Number of correct variants}}
%    \label{fig:variantResults}
%\end{figure}

\subsubsection{Robust Data- and User-adapted Visualization Output}

%data-adapted visualization
In the next experiment, we investigate how the system adapts to the resulting data.
Table~\ref{tab:comparison} shows the results of the \textit{top1/3-accuracy} in comparisons with the baseline approaches.
The topN-accuracy measures how often the desired visualization falls within the upper N predicted visualizations -- we have chosen N maximum of~3, as this number of visualizations is well presentable to the user and still usable.
We can see that our approach achieves a moderate top1-accuracy of 60.2\,\% and a very high top3-accuracy of 95.4\,\%, outperforming the baseline approaches:
\textit{CompassQL} achieves only a top1-accuracy like the majority voting in our setting -- it is even worse for the top3-accuracy.
These results also show that this issue (automated visualization adaptation) is challenging: not every adaptation method can be applied out of the box to meet user expectations.

%user-adapted visualization
In the last experiment, we investigate how the system adapts to the resulting data, \textit{additionally} taking into account the users' preferences (if any) from their intents.
The user-adapted visualization shows an improvement (64.0\,\%) for the top1-accuracy, but a small decrease (93.4\,\%) for the top3-accuracy.
An important measure here is the recall, as it indicates the fraction of the relevant types of visualization that are successfully retrieved. 
The recalls are $85.6 \pm 13.4\%$ and $92.7 \pm 10.4\%$ for the non-user-adaptive case and the user-adaptive case, respectively -- an improvement of $7.1\%$ through adaptation.

As the data is \textit{not} normally distributed (Shapiro-Wilk test, $p<.05$), we apply a non-parametric Friedman test to show the statistical differences. %for the recall
According to our results, $\chi^2(1)=10.3$, $p<.001$, there exists a statically significant difference between the top1 results of the non-user-adapted and user-adapted visualizations -- there is \textit{no} significant difference ($p=.297$) between the top3 results.

All in all, we can say that the automatic adaptation of the visual output to the data meets the expectations of the users, at least in the top3-representation -- the additional adaptation to possible user preferences from the intent can further and significantly increase the top1-representation.

\begin{table}[t]
  \centering
  \caption{Comparison against baseline approaches for data-adapting visualization}
  \begin{tabular}{rcc}
  \toprule
    Approach& top1-acc.& top3-acc.\\
  \midrule
    Majority (\textit{ZeroR})& $16.6\,\%$& $50.0\,\%$\\
    \textit{CompassQL}& $16.6\,\%$& $24.8\,\%$\\
    \textit{Insightful Assistant}& $60.2\,\%$& $95.4\,\%$\\
  \bottomrule
  \end{tabular}
  \label{tab:comparison}
\end{table}

%mmmh: table comparison against baselines, NP/P (statistical comparison), plot precision/recall for text?

\section{Conclusion}

%-contributions, why is the world better now?
In this paper, we have enabled the processing with AI-based operations and data-/user-adapted visualization in conversational agents by proposing \textit{Insightful Assistant}---a pipeline concept relying on a new operation graph representation.
We further collected a unique set of simple and complex requests to evaluate a proof-of-concept implementation.
The results show -- besides its feasibility -- an \textit{accurate} generation of the operation graph and \textit{robust} data-/user-adapted visualization output.
With \textit{Insightful Assistant}, users can trigger (easily extendable) complex (AI-based) operations that are particularly suitable for data analysis in the industrial context.

%-future work
In future work, we will extend the predefined set of functions by other AI-based algorithms to offer a wider range of data analysis.
For the top1-visualization recommendation, we plan to integrate user models that have learned from requests and feedback from the same user -- for this, we will integrate a speaker identification method to automatically distinguish the requesting users in a multi-user system.

%-acks
\section{Acknowledgements}
This \textit{collaborative} research work has been co-funded by the German Federal Ministry of Education and Research (BMBF) Software Campus (SWC) projects ``SAIS'' and ``TheNextSmartHome'' [01|S17050].

\bibliographystyle{IEEEtran}
\bibliography{insightfulAssistant}

\end{document}